%% file: main.tex
\documentclass[conference]{IEEEtran}
\IEEEoverridecommandlockouts
\usepackage{cite}
\usepackage{amsmath,amssymb,amsfonts}
\usepackage{algorithmic}
\usepackage{graphicx}
\usepackage{textcomp}
\usepackage{xcolor}
\usepackage{makecell} 
\usepackage{multirow}
\usepackage{array}
\usepackage{subcaption}
\usepackage{colortbl}
\usepackage{titling}
\def\BibTeX{{\rm B\kern-.05em{\sc i\kern-.025em b}\kern-.08em
    T\kern-.1667em\lower.7ex\hbox{E}\kern-.125emX}}
\begin{document}

\definecolor{best}{RGB}{255,180,180}      
\definecolor{second}{RGB}{255,210,160}    
\definecolor{third}{RGB}{255,240,160}     

\newcommand{\yl}[1]{{\color{cyan}\textbf{YL:} #1}}
\newcommand{\scl}[1]{{\color{brown}\textbf{SL:} #1}}

\title{TETRIS-4DGS: Efficient Post-Training Compression of \\4D Gaussian Splatting}
\date{}

\author{
Chengzhen Wu,
Kelei Liu,
Sicheng Li,
Lu Yu$^{\dagger}$,
Yiyi Liao\thanks{Corresponding author. $^{\dagger}$ Co-corresponding author.}\\
\vspace{-3pt}
Zhejiang University
}

\maketitle

\begin{abstract}
4D Gaussian Splatting (4DGS) has emerged as a powerful dynamic scene representation, supporting real-time high quality rendering but suffers from prohibitive storage overhead.
Existing methods typically attempt to solve this by treating compression as a per-scene optimization problem, limiting generalizability and applicability where low-latency compression is required. 
In this work, we leverage existing image and video codecs and propose a \textbf{generalized post-training compression framework} for 4DGS to address these issues. 
Specifically, we apply importance-based adaptive quantization to Gaussians according to their importance levels, effectively refining bitrate allocation while remaining fully compatible with existing codecs. To further improve compression efficiency, we utilize the statistical properties of 4DGS and propose motion- and importance-aware packing to reorganize attributes into codec-friendly 2D maps.
Our approach can achieve over \textbf{40× compression rate} in seconds while maintaining high quality, facilitating the practical deployment of 4DGS. 
\end{abstract}

\begin{IEEEkeywords}
4D Gaussian Splatting, 4D Gaussian Splatting Compression, Post-Training Compression 
\end{IEEEkeywords}

\section{Introduction}
\label{sec:intro}

With photorealistic and fast rendering, dynamic Gaussian Splatting is a promising pathway toward free-viewpoint video, enabling immersive visual applications. 
Existing paradigms for dynamic Gaussian Splatting can be categorized into three classes: per-frame 3DGS methods~\cite{sun20243dgstream, Jiang_2024_CVPR}, deformable Gaussians with canonical space~\cite{Wu_2024_CVPR, yang2023deformable3dgs}, and 4D Gaussian representations~\cite{yang2023gs4d, li2023spacetime}.  
Practical deployment of dynamic Gaussians requires both efficient rendering and lightweight representation. The former makes canonical-space deformable Gaussians less practical, since neural inference for deformation slows down rendering. The latter makes per-frame 3DGS less attractive due to its large storage footprint. 4D Gaussian representations—exemplified by 4DGS~\cite{yang2023gs4d}—offer a more favorable balance between rendering efficiency and model compactness.

While striking a balance,  4D Gaussian still suffer from its raw data volume, highlighting the need for effective compression. One line of work reduces the number of parameters per primitive by modifying the representation for compression, as in MEGA and CSTG. However, introducing neural networks into the representation enable memory-efficient representation but increases rendering complexity. Another line of work achieves compression by pruning low-contribution primitives, such as 4DGS-1K and the masking strategy in CSTG. These methods require costly pruning-score computation and fine-tuning.
Despite their effectiveness in terms of rate-distortion performance, these methods still treat compression as a scene-specific optimization tightly coupled with 4D Gaussian training. This tight coupling demands costly re-training when the content is not generated using the same training pipeline and becomes impractical for low-latency compression.


\begin{figure}[t]
    \centering
    \includegraphics[width=\linewidth]{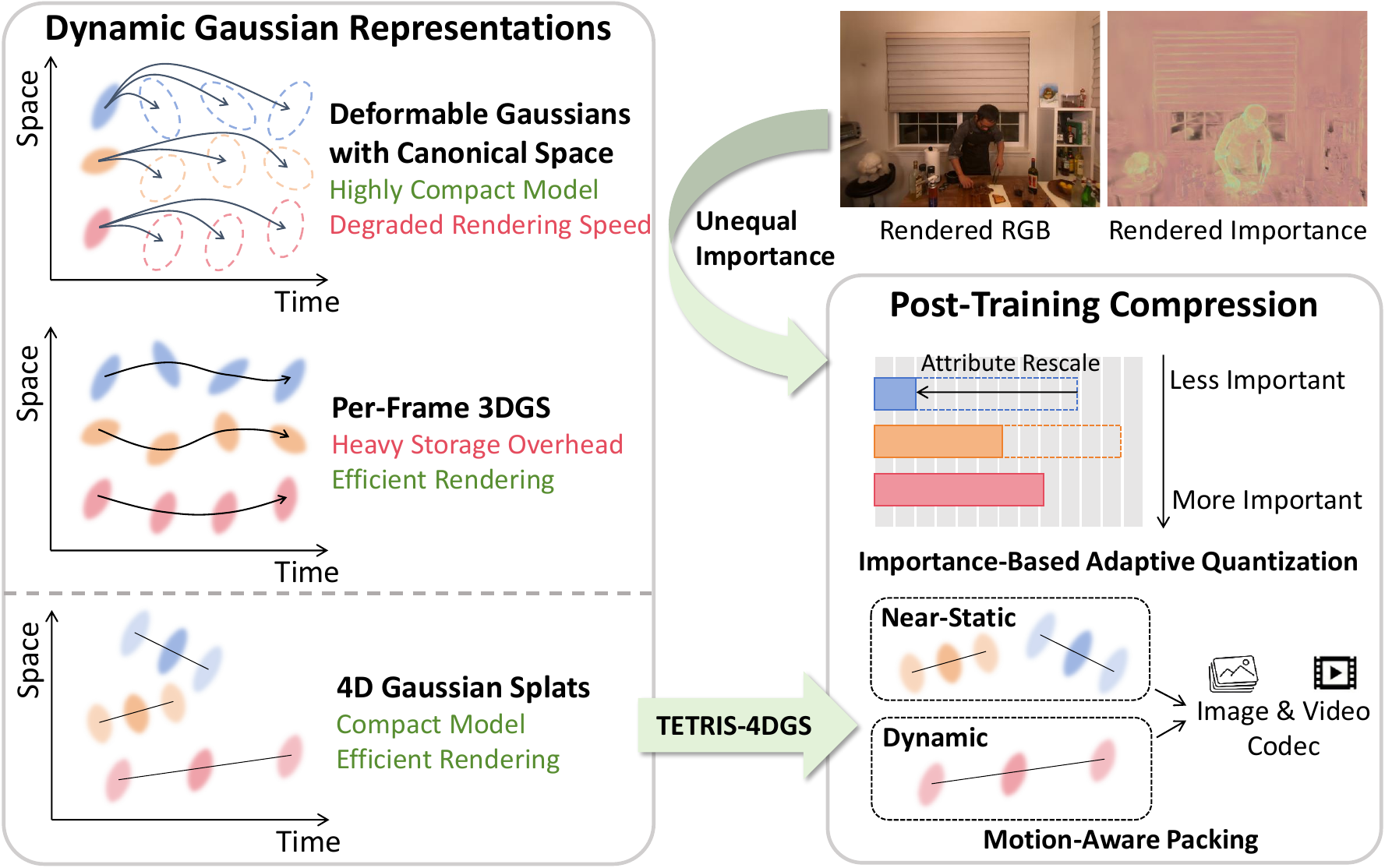}
    \caption{Targeting the 4DGS representation which balances model compactness and rendering efficiency, we exploit the unequal importance and spatial-temporal properties of 4D Gaussians and propose a post-training compression method for 4DGS that fully leverages existing image and video codecs.}
    \label{fig:teaser}
\end{figure}

We target 4DGS~\cite{yang2023gs4d} representation and aim to achieve strong rate–distortion performance for 4DGS compression with short encoding and decoding time. As shown in Fig.~\ref{fig:teaser}, we propose \textbf{TETRIS-4DGS}, an efficient post-training compression scheme for 4DGS that packs Gaussians into different blocks like Tetris.  
The key principle behind the scheme is to leverage off-the-shelf standard image and video codecs that are widely available across devices for efficient compression. 

4DGS attributes that affect primitive positions are seen as distortion-sensitive and encoded with lossless image codecs, while the remaining majority, distortion-robust attributes are encoded with lossy video codecs. To efficiently compress the latter, we introduce two techniques: 1) importance-based adaptive quantization, and 2) motion- and importance-aware packing. The former achieves adaptive bit allocation across primitives with different importance scores, without modifying the existing video codec implementation.
Unlike existing time-consuming importance calculation methods, we train a generalizable importance estimator that predicts importance scores in a feed-forward pass.
Guided by the motion statistics and importance of 4DGS, our packing method first partitions primitives into groups and then places primitives from the same group into spatially contiguous regions.
This produces codec-friendly 2D attribute maps that can be efficiently compressed by standard video codecs.

Our method achieves over \textbf{40$\times$} compression ratio while maintaining high visual quality, with only \textbf{12s} for encoding and \textbf{2s} for decoding a 300-frame dynamic scene, offering a strong rate–distortion–runtime trade-off.

\section{Related Works}

\subsection{Gaussians Primitives Pruning}


Existing methods typically address primitive redundancy by pruning Gaussians based on per-point importance metrics, which are mainly obtained in two ways:
(1) leveraging rendering kernels to calculate each primitive’s rendering contribution or uncertainty~\cite{fan2023lightgaussian, Fang2024MiniSplattingRS, HansonTuPUP3DGS, yuan20251000fps4dgaussian}; and
(2) learning per-point scores or masks through rate–distortion optimization~\cite{wang2024rdogaussian}.
While effective in improving RD performance, these methods incur substantial computational overhead and GPU memory consumption, limiting their practicality.
More recently, an MPEG contribution~\cite{xu2025gscpruning} proposed predicting per-point importance from Gaussian features using a generalized MLP, achieving competitive RD performance with reduced runtime. 
However, both this method and prior approaches categorize Gaussians into only two groups—pruned or retained—overlooking the fact that Gaussians can be more finely stratified into multiple importance levels to further exploit redundancy.

\subsection{4D Gaussian Compression}
Even though 4D Gaussian representations achieved favorable balance between model compactness and rendering efficiency, it still suffers from significant storage overhead. 
To mitigate this issue, MEGA~\cite{zhang2024mega} enlarges each Gaussian’s influence range via entropy-constrained deformation, reducing the number of primitives, and replaces spherical harmonics with a color predictor for decoding. 4DGS-1K~\cite{yuan20251000fps4dgaussian} ranks 4D Gaussians by spatial–temporal variation, pruning up to 80\% of primitives to lower storage and proceeds fine-tuning to recover quality. CSTG~\cite{Lee_2024_C3DGS} prunes Gaussians using learnable per-point masks and compresses attributes with residual vector quantization.
Though effective in terms of RD performance, these methods still tightly couple optimization with compression, lacking generalizability to pre-generated Gaussian models and making them unsuitable for low-latency compression.

\section{Method}

\begin{figure*}[t]
    \centering
    \includegraphics[width=\textwidth]{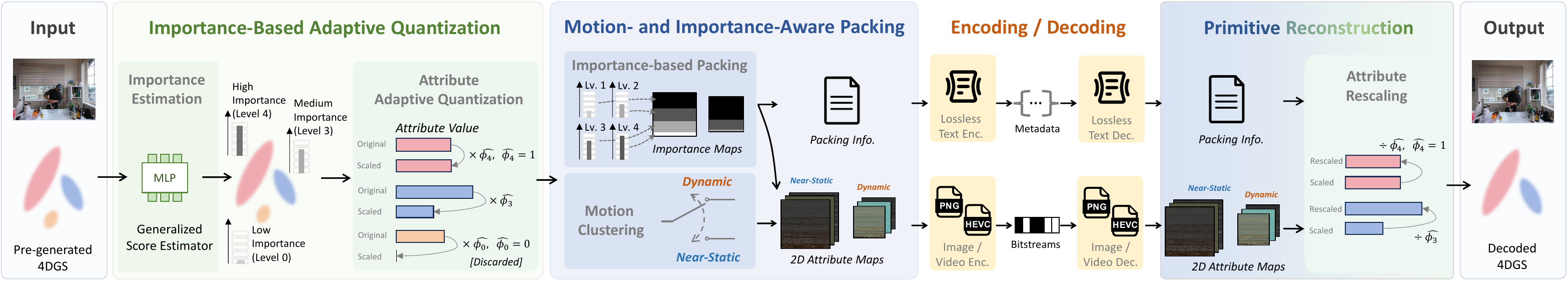}
    \caption{Overview of the proposed 4DGS generalized post-training compression framework. For encoding, a per-point importance score is first derived from the input 4DGS through the importance estimator, based on which Gaussian attributes are adaptively scaled and quantized. Attributes are then packed to 2D maps according to their importance level and motion property, making them amendable for subsequent image and video coding. During decoding, attributes are obtained from the bitstreams and rescaled to their original value, restoring the 4DGS data for further rendering.}
    \label{fig:overview}
\end{figure*}

In order to achieve strong rate–distortion–runtime trade-off on the 4DGS~\cite{yang2023gs4d} representation, we propose Importance-Based Adaptive Quantization (Sec.~\ref{sec:B}) and Motion- and Importance-Aware Packing (Sec.~\ref{sec:C}) for 4DGS compression. Fig.~\ref{fig:overview} illustrates the overall workflow of our method.

\subsection{Preliminary}
\label{sec:A}

Each 4DGS~\cite{yang2023gs4d} primitive is defined by the following attributes: (i) 4D mean vector $\mu_{4D} = (\mu_x, \mu_y, \mu_z, \mu_t) \in \mathbb{R}^4$; (ii) 4D rotation $R_{4D}$, parameterized by a pair of isotropic quaternions, $q_l \in \mathbb{R}^4$ and $q_r \in \mathbb{R}^4$; (iii) 4D scaling factor $s_{4D} = (s_x, s_y, s_z, s_t) \in \mathbb{R}^4$; (iv) spatial and temporal spherical harmonics (SH) $sh \in \mathbb{R}^{3(k_v+1)^2(k_t+1)}$ with view degrees of freedom $k_v$ and time degrees of freedom $k_t$; (v) opacity $o \in [0,1]$. Given the above attributes, the covariance matrix $\Sigma_{4D}$ of a 4DGS can be constructed as:

\begin{equation}
\Sigma_{4D} = R_{4D}S_{4D}S_{4D}^T R_{4D}^T = 
\begin{pmatrix}
U & V \\
V^T & W
\end{pmatrix},
\label{eq:cov4d}
\end{equation}
where $U \in \mathbb{R}^{3 \times 3}$, $V \in \mathbb{R}^{3 \times 1}$, and $W \in \mathbb{R}^{1 \times 1}$ represent the decomposed sub-matrices of the covariance.

For rendering, 4D Gaussians are projected into 3D space, where $\Sigma_{3D} = U - \frac{VV^T}{W}$ denotes the 3D covariance matrix and $\sigma(t) = e^{-\frac{(t-\mu_t)^2}{2W}}$ models the temporal decay of opacity. Furthermore, the 3D centers of Gaussians are defined as $\mu_{3D}(t) = \mu_{3D} + (t-\mu_t)\frac{V}{W}$, approximating Gaussian trajectories as uniform linear motions. Following the projection of 4DGS into 3D space, differentiable rasterization is applied to the projected 3D Gaussians to synthesize the final image.

According to the above formulation, the spherical harmonics and opacity of 4DGS are distortion-robust for they do not affect the Gaussian position, whereas the remaining attributes influence their positions to some extent. Our work focuses on efficiently compressing the former group—which constitutes the majority of attribute channels—using lossy video codecs, as these attributes contain greater redundancy.

\subsection{Importance-Based Adaptive Quantization}
\label{sec:B}


While current pruning-based methods reduce the primitive redundancy among 4DGS, they treat all surviving Gaussians equally. This binary approach is inefficient for primitives of borderline importance: pruning them degrades quality, yet storing them at full precision wastes bits. These primitives should be treated in a more delicate way, for which we propose importance-based adaptive quantization.

\noindent \textbf{\textit{Importance-Based Adaptive Quantization for Attributes: }} 
We assign each Gaussian attribute a specific bit width and then uniformly quantize it using its per-channel minimum and maximum values, where we propose an importance-guided attribute rescaling. Specifically, each Gaussian is associated with an importance score $\phi \in (0, 1)$, predicted by a generalized importance score estimator, which is described in the following paragraph. We then divide the importance score $\phi$ into $L$ importance levels $l$:
\begin{equation}
l(\phi) =
\begin{cases}
0, & \phi \in (0, \phi_{\mathrm{thres}}), \\
1 + \left\lfloor (L-1) \cdot \dfrac{\phi - \phi_{\mathrm{thres}}}{1 - \phi_{\mathrm{thres}}} \right\rfloor, 
& \phi \in [\phi_{\mathrm{thres}}, 1),
\end{cases}
\end{equation}
where $l \in \{0,1,2,...,L-1\}$ denotes the assigned importance level and $\phi_{\mathrm{thres}}$ denotes a preset score threshold which will be explained later.
With the importance level of each primitive determined, we define the corresponding attribute rescale ratio $\hat{\phi}_L$ as:
\begin{equation}
\hat{\phi}_l =
\begin{cases}
0, & l = 0, \\[6pt]
\phi_l^{\max}, & l \in \{1,2,...,L-2\}, \\[6pt]
1, & l = L-1, 
\end{cases}
\end{equation}
where $\phi_l^{\max}$ denotes the maximum score value of level $l$ importance scores. 

Next, the distortion-robust Gaussian attributes $A$ are multiplied by their respective rescale ratios $\hat{\phi}_l$ before quantization, forming adjusted attribute value $A'$:
$A' = A \cdot \hat{\phi}_l$.
%
During decoding, the original attribute values will be restored by dividing their own rescale ratio:
$\hat A = \hat A' / \hat{\phi}_l$.
%
%
Through this process, while Gaussian attributes share the same per-channel quantization range and bit width, the value ranges of attributes from less important Gaussians are narrowed down, equivalent to being quantized at a lower precision. When combined with subsequent video coding, this strategy plays a role analogous to the rate–distortion optimization process in video codecs, while requiring no major modifications to existing codec architectures.

It is worth noting that Gaussians assigned to the lowest importance level are permanently discarded, as their attributes are multiplied by zero and can no longer be restored. This process is equivalent to pruning and can be controlled by adjusting a predefined lowest importance ratio, which in turn determines the corresponding score threshold $\phi_{\mathrm{thres}}$. 

%
\noindent \textbf{\textit{Generalized Importance Score Estimator:}} 
%
We now explain how to train a generalized score estimator to predict the aforementioned importance score $\phi$. The score estimator we propose takes features that describe each primitive's spatial and temporal properties as input, and outputs a per-point predicted importance score. 

We extract seven feature types from each 4D primitive: the average KNN distance $d$, computed at the midpoint of its uniform linear motion; the 4D Gaussian scales $(s_x, s_y, s_z, s_t)$; the 4D volume $V = s_x s_y s_z s_t$; the opacity $\alpha$; the color anisotropy $A$, defined as the ratio of color variance to squared mean over $M$ sampled directions on the bounding sphere; the motion magnitude $v$; and the lifespan $T$, measured as the temporal interval where the opacity decay $\sigma(t)$ exceeds a threshold $\tau$.
Given the resulting 10-channel feature set $\{d, s_x, s_y, s_z, s_t, V, \alpha, A, v, T\}$ and the $K$ nearest neighbors of each Gaussian, we standardize each feature both globally across all features and locally within its neighborhood, and concatenate the resulting features to feed into the score estimator.
With the input of the score estimator prepared, we then obtain the corresponding supervision by training per-point importance scores in a rate-distortion optimization manner on the training dataset. Specifically, we assign a sigmoid function constrained importance score $\phi^{s} \in (0, 1)$ to each Gaussian. Based on the score, we proceed simulation of importance-based adaptive quantization, where the quantization step size $\Delta$ is defined as:
\begin{equation}
\Delta(\phi^{s}) = \frac{\Delta_0}{\phi^{s}}
\end{equation}
Here, $\Delta_0$ denotes the base quantization step size. Through this design, Gaussians have their quantization step size adjusted according to their respective importance score, effectively simulating the possible perturbation caused by adaptive quantization. Additionally, the least important Gaussian attributes will all be quantized to zero to faithfully simulate the effect of discarding them.
Quantization is simulated using a straight-through estimator (STE) to enable gradient propagation, as quantization is inherently a discrete operation.


For optimization, we introduce a rendering quality loss to preserve representation accuracy, as defined in Eq.~\eqref{eq:render_loss}, and a sparsity loss to encourage smaller importance scores, defined in Eq.~\eqref{eq:sparcity_loss}. The two terms formulate a rate–distortion joint training objective, enabling to learn rate–distortion–efficient importance scores as supervisions.
\begin{equation}
\mathcal{L}_{\mathrm{render}} =
\lVert \mathbf{I} - \mathbf{I}_{\mathrm{gt}} \rVert_1
+ \lambda_{\mathrm{dssim}}
\big( 1 - \mathrm{SSIM}(\mathbf{I}, \mathbf{I}_{\mathrm{gt}}) \big).
\label{eq:render_loss}
\end{equation}
\begin{equation}
\mathcal{L}_{\mathrm{sparcity}}
= \frac{1}{N} \sum_{i=1}^{N} \phi_i^{s},
\label{eq:sparcity_loss}
\end{equation}
With both the input features and supervision scores prepared, we train a lightweight MLP as a score estimator on the training scenes. Once trained, the estimator generalizes well to unseen scenes, producing accurate and robust importance scores efficiently to guide importance-based adaptive quantization. 


\subsection{Motion- and Importance-Aware Packing}
\label{sec:C}

Based on 4D primitives' predicted importance and their spatial-temporal properties, we propose Importance-based Packing and Motion-Aware Packing to make 4DGS attributes more amendable for existing codecs. 

\noindent\textbf{\textit{Importance-based Packing:}} 
During importance-based adaptive quantization, the distortion-robust Gaussian attributes are rescaled. 
Therefore, attributes at different importance levels exhibit distinct statistical distributions. This necessitates separate handling when reorganizing 4DGS attributes into 2D attribute maps for image and video coding.
Specifically, we apply Morton sorting to the trajectory centers $\overline{\boldsymbol{\mu}}_{3D}$ of 4DGS within each importance level, evaluated at the time of maximum opacity. After obtaining the 1D Morton order for each level, we concatenate them into a unified 1D ordering. Based on this ordering, we reorganize 4DGS attributes into a 2D pixel grid using row-major layout, which balances packing efficiency and spatial locality.
Finally, the reorganized 2D maps of distortion-sensitive attributes are compressed using PNG. For distortion-robust attributes, higher-order SH maps are concatenated into a multi-frame video, while opacity and zeroth-order SH maps form single-frame videos; both are compressed using HEVC via FFmpeg.
\\
\hspace*{\parindent}Through this design, Gaussians of different importance levels are placed in separate regions, allowing less important attributes' region to be smoother and easier for video codes to encode. It is worth noting that the importance map, doesn't have to be encoded and transmitted for restoring the rescaled attribute values. Instead, thanks to the reorganization method described above, it suffices to store and transmit only the coordinates of the $L-2$ pixels that mark the boundaries between the $L-1$ regions in the importance map, along with the $L-1$ regions corresponding rescale ratio $\hat{\phi}_l$.

\begin{figure}[t]
    \centering
    \includegraphics[width=0.8\linewidth]{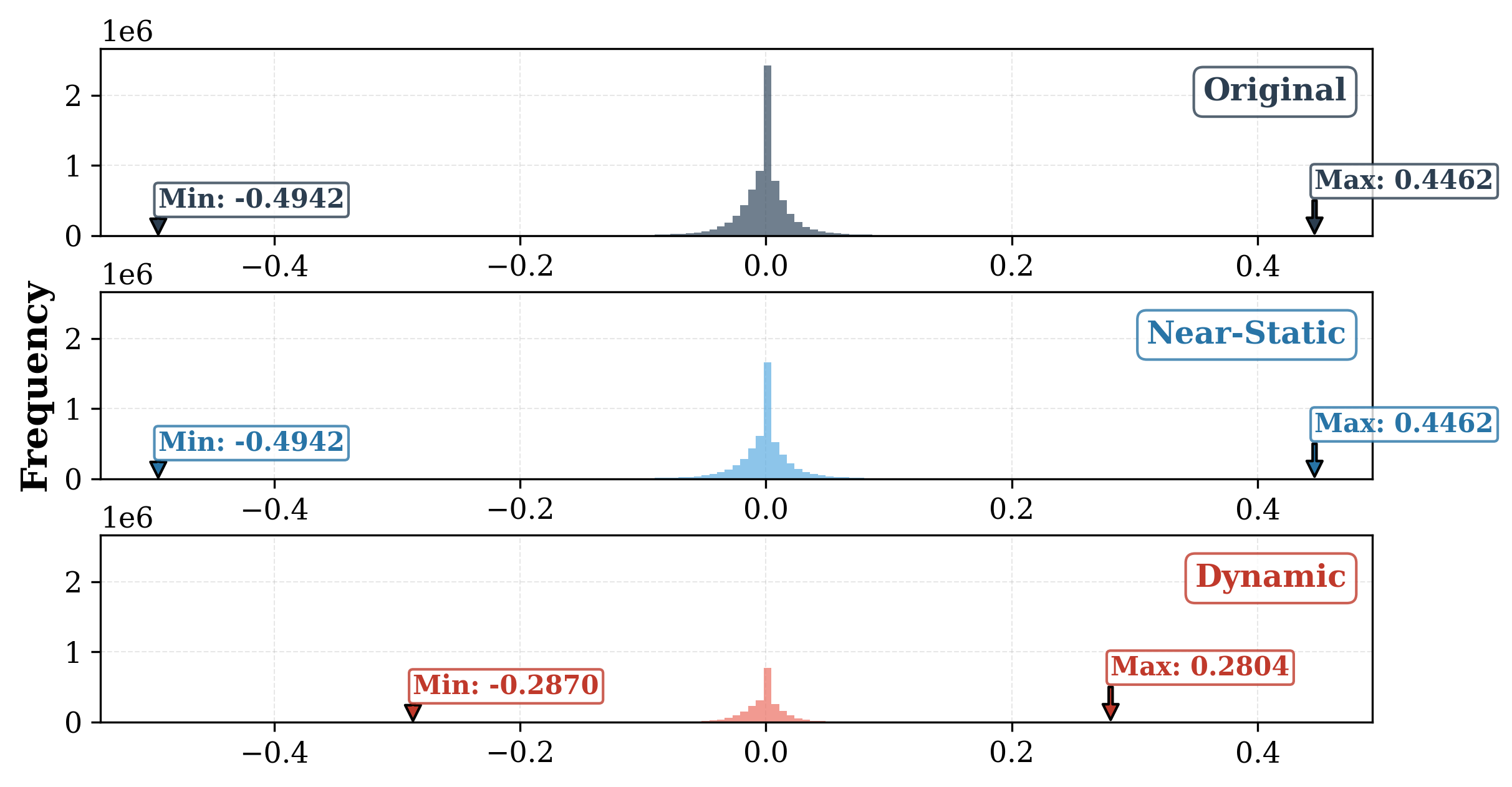}
    \caption{Example histograms of one sh$_N$ channel's distribution.}
    \label{fig:shN_distribution_change}
\end{figure}

\noindent\textbf{\textit{Motion-Aware Packing:}} 
To exploit the spatial-temporal characteristics of 4DGS for compression, we analyze the motion statistics of 4D primitives and observe a Laplacian-like distribution: most Gaussians exhibit near-zero motion, while only a small fraction move rapidly. Further analysis reveals a consistent difference in attribute distributions between these two groups. For example, Fig.~\ref{fig:shN_distribution_change} shows that the high-order spherical harmonics coefficients (sh$_N$) of dynamic Gaussians lie in a significantly narrower range than those of near-static ones. This is because dynamic Gaussians are observed under fewer stable view directions, which suppresses high-frequency angular variations. This observation generalizes to other attribute channels like zero-order spherical harmonics and the imaginary parts of both quaternions.
\\
\hspace*{\parindent}Based on this observation, we partition 4DGS into \textit{near-static Gaussians}, whose motion magnitudes along all axes are below a threshold $v_{\text{th}}$, and \textit{dynamic Gaussians}, where at least one axis exceeds this threshold. 
We then apply motion-aware packing by processing these two groups independently. Each group is reorganized into separate attribute maps and quantized using group-specific minimum and maximum values, while sharing the same bit width. As dynamic Gaussians occupy a smaller value range, this strategy effectively increases quantization precision, leading to improved rate-distortion performance without additional cost.

\section{Experiments}

\begin{figure*}[t]
    \centering
    \includegraphics[width=0.8\linewidth]{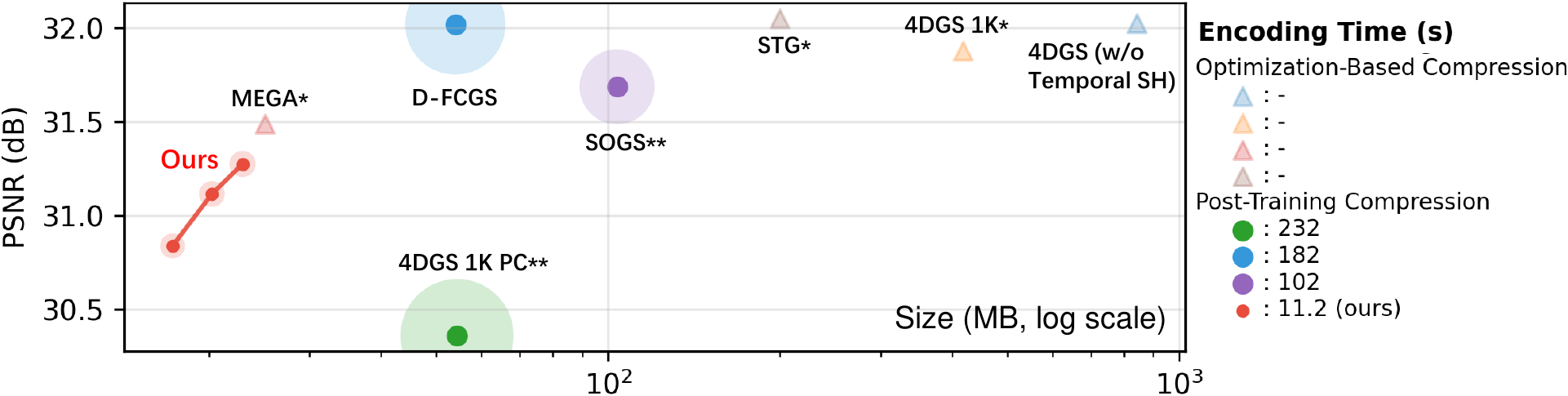}
    \caption{RD Performance comparison conducted on the Neural 3D Video Dataset. Methods marked with * and triangle are optimization-based methods; methods marked with ** indicate our reimplementations of existing methods adapted for post-training 4DGS compression. 4DGS (w/o Temporal SH) serves as the starting point of our compression scheme. We further report the encoding time for our methods and other post-training approaches, represented by the size of the outer circle.}
    \label{fig:RD_curve}
\end{figure*}

\begin{figure}[t]
    \centering
    \includegraphics[width=0.8\linewidth]{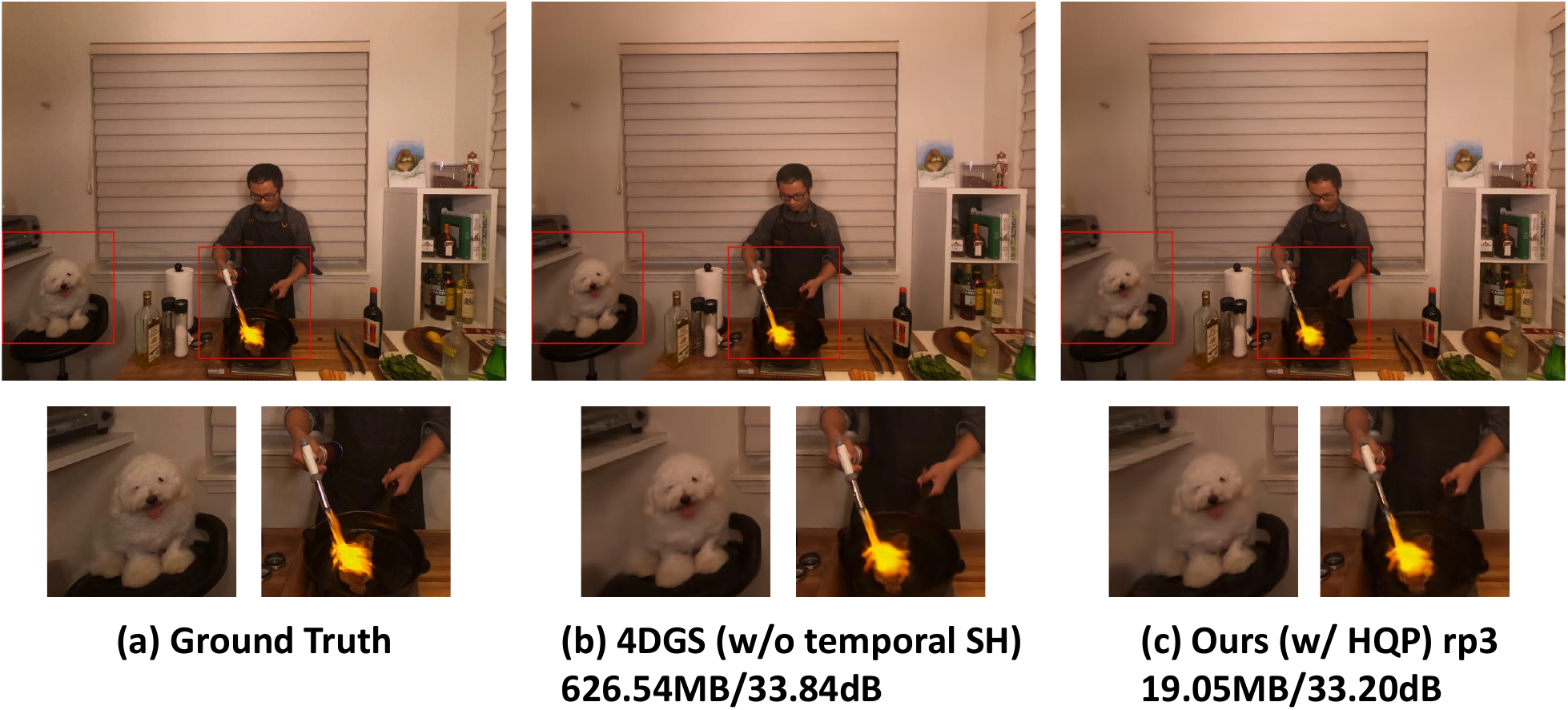}
    \caption{Qualitative comparisons on Neural 3D Video dataset.}
    \label{fig:qualitative_comparison}
\end{figure}

\input{table/comparison_all_in_one}

\subsection{Experiment Settings}

\noindent\textbf{\textit{Datasets:}} 
For training, we select three representative scenes in the MeetRoom \cite{li2022streaming} dataset to reconstruct 4DGS and derive per-point importance scores, serving as the supervision targets for the importance score estimator. For evaluation,  we conduct experiments on the Neural 3D Video \cite{Li_2022_CVPR} dataset, using the initial 300 frames of each scene for all evaluations.


\noindent\textbf{\textit{Implementation Details:}} 
We adopt the official implementation of 4DGS \cite{yang2023gs4d} to train 4DGS models on the two datasets. Notably, we do not employ the extremely redundant temporal spherical harmonics proposed in the original work. Instead, we use 48 channel spherical harmonics following Inria 3DGS~\cite{kerbl3Dgaussians} and are able to reproduce the reported rendering quality.
After obtaining the per-point importance scores as supervision, we train an MLP-based score estimator, whose layer dimensions are [256, 128, 64, 64, 32], to regress importance values.
%
We set the number of importance levels $L$ to 5 in our experiments. 
\noindent\textbf{\textit{Baselines:}} Post-training compression for 4DGS remains largely unexplored, and no dedicated baselines currently exist. To enable a comprehensive evaluation, we compare our method against two categories of approaches.
\textit{(1) Post-training compression methods.}
We include D-FCGS~\cite{zhang2025d}, a post-training compression framework designed for dynamic Gaussian representations that operates on per-frame 3DGS sequences rather than 4DGS. We further adapt representative 3D or 4D compression methods to the post-training setting to compress the same 4DGS data for fair comparison. Specifically, we consider 4DGS-1K PC, adapted from 4DGS-1K~\cite{yuan20251000fps4dgaussian} by removing the post-pruning fine-tuning stage, and SOGS~\cite{morgenstern2024compact}, which we extend to 4DGS by sorting 4D Gaussians into 2D image layouts using its proposed sorting strategy.
\textit{(2) Optimization-based compression methods.}
We also compare with approaches that integrate compression into the reconstruction pipeline, including MEGA~\cite{zhang2024mega}, 4DGS-1K~\cite{yuan20251000fps4dgaussian}, Spacetime Gaussians~\cite{li2023spacetime}, and the original 4DGS~\cite{yang2023gs4d} as a reference. These methods jointly optimize reconstruction and compression through end-to-end training or extensive per-scene fine-tuning, whereas our approach focuses on post-training compression without retraining.
%


\subsection{Comparison with State-of-the-Art Methods}

Table~\ref{tab:comparison_all_in_one} reports rate–distortion (RD) performance in terms of PSNR, SSIM, and LPIPS, together with rendering FPS and encoding/decoding time on the Neural 3D Video dataset. Fig.~\ref{fig:RD_curve} further shows PSNR–bitrate and encoding-time comparisons. Our method achieves superior RD performance over most post-training compression approaches while requiring substantially less encoding time. Compared with state-of-the-art optimization-based methods such as MEGA, our approach attains comparable RD performance while delivering significantly higher rendering FPS, as it avoids feature decoding during rendering. Moreover, our method is more flexible across devices with different GPU memory budgets, since the MLP-based importance prediction does not require processing all Gaussians simultaneously, unlike rendering-based scoring methods. Qualitative results in Fig.~\ref{fig:qualitative_comparison} further demonstrate that our method preserves visual quality comparable to the 4DGS baseline while achieving a $33\times$ compression ratio.



\subsection{Ablation Study}

\begin{figure}[t]
    \centering
    \begin{subfigure}{0.38\linewidth}
        \centering
        \includegraphics[width=\linewidth]{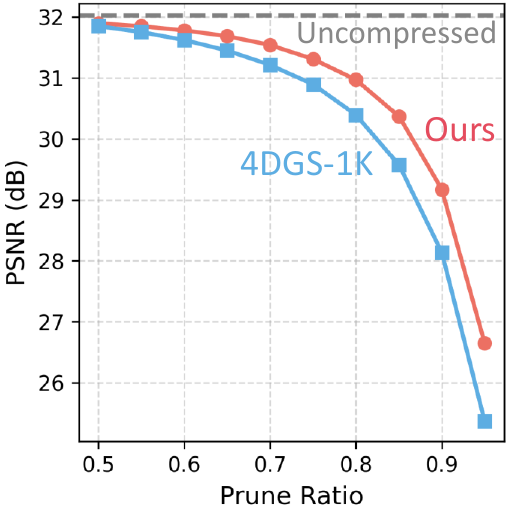}
        \caption{PSNR vs. Prune ratio.}
        \label{fig:PSNR_vs_prune_ratio}
    \end{subfigure}
    \hfill
    \begin{subfigure}{0.58\linewidth}
        \centering
        \includegraphics[width=\linewidth]{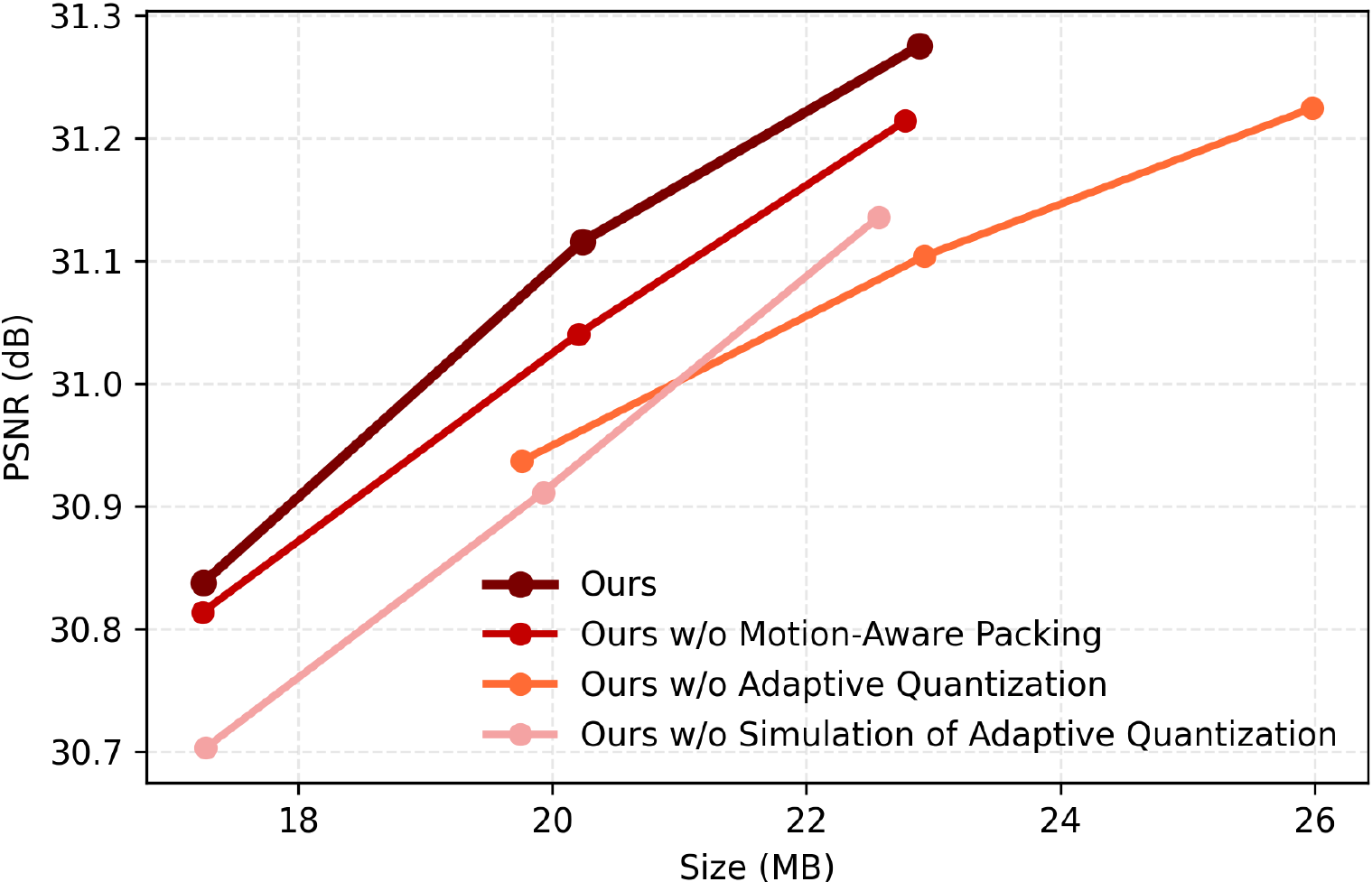}
        \caption{Ablation RD curves.}
        \label{fig:SoftPruning_Ablation}
    \end{subfigure}
    \caption{Ablation experiments on Neural 3D Video dataset.}
    \label{fig:PSNR_and_Ablation}
\end{figure}

To assess the effectiveness of our method, we perform ablation studies on the Neural 3D Video dataset focusing on the following three components:

\noindent\textbf{\textit{Generalized Score Estimator:}} 
We assess our score estimator based on pruning guidance effectiveness, GPU memory usage, and computational time. 
We first compare our generalized importance score with the spatiotemporal variation score proposed in 4DGS-1K, which requires rendering all training views and computing importance from opacity and lifespan. Fig.~\ref{fig:PSNR_vs_prune_ratio} shows the PSNR–pruning ratio curves, where our approach achieves better performance using an efficient, generalizable importance estimator. Note that here we set our importance level number $L$ to 2 for fair comparison with 4DGS-1K.

%
Next, we evaluate the flexibility of our method in terms of GPU memory usage and computational efficiency in Table~\ref{tab:memory_runtime}, using the Coffee Martini scene with approximately 4.5 million primitives. While 4DGS-1K incurs high GPU memory consumption by loading all active primitives into the rendering pipeline simultaneously, our method allows for flexibly adjusting the input feature batch size to match available device memory. In terms of runtime, our approach completes in 4 seconds, whereas 4DGS-1K requires approximately 3 minutes due to the need to render all training views.

\input{table/GPU_flexibility}

\noindent\textbf{\textit{Importance-Based Adaptive Quantization for Attributes:}} 
In this section, we evaluate the rate-distortion performance improvements achieved by the proposed importance-based adaptive quantization, as well as the effect of simulating this strategy during the supervision training of the generalized score estimator. The RD comparisons are presented in Fig.~\ref{fig:SoftPruning_Ablation}. To assess the effectiveness of adaptive quantization, we compare two settings with different numbers of importance levels: $L=2$ (Ours w/o Adaptive Quantization) and $L=5$ (Ours). Our method achieves an approximate 12\% bitrate reduction at the highest bitrate point with no PSNR drop, indicating that it effectively removes redundancy among Gaussians while maintaining quality. 

Furthermore, we analyze the contribution of adaptive quantization simulation during supervision training. When adaptive quantization is not simulated during training (i.e., only pruning is simulated with $L=2$), but adaptive quantization is still applied during post-training processing with $L=5$, a noticeable degradation in rendering quality is observed. This behavior suggests that simulating adaptive quantization during training leads to importance scores that are more robust to adaptive quantization at compression time, thereby alleviating the quality loss introduced by this process.

\noindent\textbf{\textit{Motion-Aware Packing:}} 
Our separate packing strategy maintains the same quantization bit width for both dynamic and near-static Gaussians. As a result, dynamic Gaussian attributes are quantized with higher precision without introducing additional bit width due to their reduced value range. We further assign a higher QP to the high-order spherical harmonics of dynamic Gaussians when encoded with video codecs to optimize bitrate allocation. This leads to a clear improvement in RD performance, as demonstrated in Fig.~\ref{fig:SoftPruning_Ablation}.

\section{Conclusion}


This paper introduces a generalized post-training compression framework for 4DGS that achieves competitive rate-distortion performance with minimal encoding and decoding time. 
Through importance-based adaptive quantization and motion- and importance-aware packing, guided by a lightweight and generalizable importance estimator, we effectively exploit redundancy in the 4DGS representation. Experimental results demonstrate that our approach delivers a strong rate–distortion trade-off with very short runtime and high flexibility across GPU memory constraints.

\section*{Acknowledgment}
This work was supported by the National Natural Science Foundation of China under Grant 62441223 and 62571485.

\bibliographystyle{IEEEbib}
\bibliography{icme2026references}

\end{document}

%% file: table/comparison_all_in_one.tex
\setlength{\tabcolsep}{8pt}
\begin{table*}[t]
\refstepcounter{table} 

{\normalsize
\textbf{Table \thetable:} Quantitative comparison on Neural 3D Video dataset. ``Size" refers to the total model size for 300 frames. Methods marked with * indicate that the results are directly taken from the original paper. Methods marked with ** indicate results obtained from our reimplementation adapted for 4DGS.
}
\vspace{1mm} 


\centering
\label{tab:comparison_all_in_one}
\begin{tabular}{lccccccc}

\hline
\textbf{Method} & \textbf{PSNR ↑} & \textbf{LPIPS ↓} & \textbf{SSIM ↑} & \textbf{Size~(MB) ↓} & \textbf{FPS ↑} & \textbf{Enc. Time~(s) ↓} & \textbf{Dec. Time~(s) ↓} \\


\hline
\textit{Optimization-Based Compression}& & & & & & &\\
MEGA* & 31.49 & 0.0568 & 0.9420 & 25.05 & 77 & - & - \\
4DGS 1K* & 31.88 & 0.0530 & 0.9440 & 418.00 & 805 & - & - \\
STG* & 32.05 & 0.0440 & 0.9480 & 200.00 & 140 & - & - \\
4DGS* & 32.01 & 0.0550 & 0.9720 & - & 114 & - & - \\
4DGS (w/o temporal SH) & 32.03 & 0.0924 & 0.9453 & 843.29 & 95 & - & - \\
\hline
\textit{Post-Training Compression}& & & & & & &\\
D-FCGS* & \cellcolor{best}32.02 & - & \cellcolor{best}0.9510 & 54.00 & \cellcolor{best}215 & 182 & 215.28 \\
SOGS for 4DGS** & \cellcolor{third}31.69 & \cellcolor{best}0.0988 & \cellcolor{second}0.9399 & 103.64 & 95 & 102 & 2.33 \\
4DGS 1K PC** & 30.36 & 0.1362 & 0.9265 & 54.39 & - & 232 & \cellcolor{best}0.11 \\
Ours (Low) & 30.83 & 0.1175 & 0.9309 & \cellcolor{best}17.25 & \cellcolor{second}134 & \cellcolor{best}11.20 & \cellcolor{second}1.90 \\
Ours (Medium) & 31.11 & 0.1121 & 0.9331 & \cellcolor{second}20.23 & \cellcolor{third}130 & \cellcolor{second}11.99 & \cellcolor{third}2.10 \\
Ours (High) & 31.27 & \cellcolor{third}0.1084 & 0.9344 & \cellcolor{third}22.89 & 128 & \cellcolor{third}12.48 & 2.27 \\
Ours (Very High) & \cellcolor{second}31.71 & \cellcolor{second}0.0994 & \cellcolor{third}0.9395 & 36.33 & 128 & 12.51 & 2.28 \\


\hline
\end{tabular}

\end{table*}

%% file: table/GPU_flexibility.tex
\begin{table}[t]
\refstepcounter{table} 

{\normalsize
\textbf{Table \thetable:} Peak GPU memory usage and runtime comparison. For our method, we report the feature extraction time and the MLP inference time, respectively. The number in () denotes the batch size of input features to our score estimator.
}
\vspace{1mm} 

\centering

\setlength{\tabcolsep}{5pt} 
\renewcommand{\arraystretch}{1.05}

\begin{tabular}{c c c c}
\hline
\makecell{\textbf{Device Memory}\\\textbf{Limit}} &
\textbf{Method} &
\makecell{\textbf{Peak Memory}\\\textbf{(GB)}} &
\textbf{Runtime (s)}\\
\hline

\multirow[c]{2}{*}{8 GB} & Ours (2.5M)   & 6.21 & $3.91 + 0.04$ \\
                         & 4DGS-1K & 5.92 & 184 \\
\hline


\multirow[c]{2}{*}{4 GB} & Ours (1.25M)  & 3.84 & $3.91 + 0.06$ \\
                         & 4DGS-1K & OOM  & OOM \\
\hline

\multirow[c]{2}{*}{2 GB} & Ours (0.15M)  & 1.75 & $3.91 + 0.13$ \\
                         & 4DGS-1K & OOM  & OOM \\
\hline
\end{tabular}

\label{tab:memory_runtime}
\end{table}